\documentclass{article}
\usepackage[preprint,nonatbib]{neurips_2026}
\usepackage[T1]{fontenc}
\usepackage[utf8]{inputenc}
\usepackage{microtype}
\usepackage{amsmath,amssymb,amsthm,mathtools,bm}
\usepackage{booktabs,multirow,array}
\usepackage{enumitem}
\usepackage{xcolor}
\usepackage{graphicx}
\usepackage{url}
\usepackage{hyperref}
\usepackage[nameinlink,noabbrev]{cleveref}
\hypersetup{colorlinks=true,citecolor=blue,linkcolor=blue,urlcolor=blue}

\newtheorem{theorem}{Theorem}
\newtheorem{lemma}{Lemma}
\newtheorem{proposition}{Proposition}
\newtheorem{corollary}{Corollary}

\crefname{theorem}{Theorem}{Theorems}
\Crefname{theorem}{Theorem}{Theorems}
\crefname{lemma}{Lemma}{Lemmas}
\Crefname{lemma}{Lemma}{Lemmas}
\crefname{proposition}{Proposition}{Propositions}
\Crefname{proposition}{Proposition}{Propositions}
\crefname{corollary}{Corollary}{Corollaries}
\Crefname{corollary}{Corollary}{Corollaries}
\crefname{definition}{Definition}{Definitions}
\Crefname{definition}{Definition}{Definitions}

\DeclareMathOperator{\KL}{KL}

\DeclareMathOperator{\diag}{diag}

\DeclareMathOperator{\Beta}{Beta}

\DeclareMathOperator{\TV}{TV}

\newcommand{\E}{\mathbb{E}}

\newcommand{\given}{\,|\,}

\newcommand{\PE}{\mathrm{PE}}
\newcommand{\Acal}{\mathcal{A}}

\providecommand{\answerYes}[1]{\textbf{Yes}}
\providecommand{\answerNo}[1]{\textbf{No}}
\providecommand{\answerNA}[1]{\textbf{N/A}}

\title{LLMs are Bayesian in Expectation, Not Realization}
\author{%
  Leon Chlon\\
  Hassana Labs\\
  \texttt{leo@hassana.io}
  \And
  Zein Khamis\\
  Hassana Labs
  \And
  Fatima Sheaib\\
  Hassana Labs
  \AND
  Maggie Chlon\\
  Hassana Labs
  \And
  Mahdi El Zein\\
  Hassana Labs
  \And
  MarcAntonio M.~Awada\\
  Harvard University\\
  \texttt{mawada@hbs.edu}
}
\date{}

\begin{document}
\maketitle

\begin{abstract}
Bayesian accounts of in-context learning appear to face a direct objection: exact posterior predictives for exchangeable data are invariant to task-preserving order, while transformers often change next-token probabilities when the same examples are serialized differently. We show that this objection targets a structural invariant rather than the operational quantity used to score online prediction. For any Bayesian predictive reference, excess prequential code length is exactly cumulative predictive KL. For unordered support sets that must be serialized, the expected regret of a single admissible ordering decomposes into the regret of the order-averaged predictor plus an order-averaging gain. Exchangeability violations are therefore not binary refutations; they are priced by log loss. We instantiate the theory with KT/Dirichlet finite-alphabet prediction, candidate-event decompositions, safe-code floors, and coarsened Bayesian linear-regression (BLR) predictive distributions. On Qwen2.5-7B/14B, floored candidate distributions at support $256$ have one-step excess code lengths of $0.020/0.011$ bits for Bernoulli prediction and $0.039/0.022$ bits for four-way categorical prediction, with candidate mass above $0.999$; coarsened BLR continuations increasingly match the posterior-predictive digit distribution as support grows. A frequentist plug-in baseline sharpens the reading: the predictive distributions sit closer to the Bayesian posterior predictive than to the maximum-likelihood plug-in, by a margin largest at small support---where the plug-in is degenerate---and vanishing as the references converge. Position interventions and a from-scratch encoding ablation separate task-preserving from semantic order and localize it to the positional encoding; activation patching tests whether decoded sufficient statistics causally affect predictions; and permutation mixtures quantify the log-loss cost of arbitrary orderings downstream. The result is a quantitative reconciliation: transformers need not realize exchangeable posterior predictives for every serialization to be Bayes-competitive prequential predictors.
\end{abstract}

\section{Introduction}
\label{sec:intro}

Bayesian interpretations of in-context learning give few-shot prompting a clean semantics: the support examples identify a latent task, and the continuation is the posterior predictive under that task \cite{xie2021explanation,muller2021transformers,bai2023transformers,muller2024bayespower}. This view predicts a concrete object: the conditional distribution for the next token. It also appears to collide with a concrete fact about transformers. A Bayesian posterior predictive for exchangeable observations is invariant to task-preserving reordering; an autoregressive transformer reads a sequence with positions.

Recent martingale and exchangeability diagnostics make this collision explicit. Pretrained language models do not, in general, satisfy the identities of exact Bayesian prediction on exchangeable data: permuting an unordered support set can change next-token probabilities, and posterior-predictive martingale tests can fail \cite{falck2024martingale,jesson2024martingaleicl}. These results are important. They rule out the strongest slogan version of Bayesian ICL: a transformer is not an exact exchangeable posterior-predictive distribution for every serialization of the same support set.

They do not rule out the operational claim that matters for online prediction. A sequence predictor is scored by prequential log loss,
\begin{equation}
\label{eq:preq-loss}
L_n(Q,y_{1:n})=\sum_{t=1}^n -\log q_t(y_t\given y_{<t}),
\end{equation}
where $q_t(\cdot\given y_{<t})$ is the conditional distribution actually used at prefix $t$. This score does not directly penalize failure of invariance at unused serializations. It penalizes probability assigned to the next observation. If the model's conditional distributions stay close to a Bayesian reference in predictive KL along the evaluated path, the cumulative code length can remain close to the Bayesian code even when exact exchangeability fails.

The central contribution is therefore a change of estimand. Structural exchangeability is a property of an exact posterior-predictive law. Prequential regret is the operational quantity paid by a sequence predictor. The bridge between them is predictive KL. We prove this first as an exact prequential comparison identity. We then show how arbitrary serialization of an unordered information state contributes an additional, measurable order-averaging gain. This turns exchangeability from a yes/no test into a code-length budget.

\paragraph{Thesis.}
Exchangeability failures refute exact posterior-predictive equivalence. They do not, by themselves, refute Bayes-competitive prequential prediction. The operational question is how much cumulative predictive KL the violations spend.

\paragraph{Contributions.}
First, we give a robust formalization of the evaluation target: excess prequential code length against any Bayesian reference is cumulative predictive KL. Second, we introduce a task-preserving ordering decomposition: the regret of a single serialization equals the regret of the order-averaged predictor plus a nonnegative order-averaging gain. Third, we connect language-model scoring to native probabilities through a candidate-event decomposition and safe-code floors with explicit overhead. Fourth, we audit pretrained transformers in closed-form Bayesian settings, coarsened BLR continuations, position interventions, causal statistic-use interventions, and downstream order-averaging experiments. Fifth, we add a non-Bayesian baseline---the maximum-likelihood and Laplace plug-in predictors---and report the model-to-Bayes minus model-to-plug-in predictive code length, which tests whether a frequentist sequential predictor accounts for the same conditionals. The experiments are not presented as evidence that transformers are literal posterior samplers. They estimate the code-length terms that determine whether order sensitivity is operationally costly.

\paragraph{Relation to prior work.}
Bayesian and latent-task accounts explain ICL as approximate posterior prediction \cite{xie2021explanation,muller2021transformers,muller2024bayespower}; algorithmic accounts show that transformers can implement learned predictors, gradient-descent-like updates, or algorithm selection in context \cite{garg2022can,vonOswald2023transformers,akyurek2023algorithmicl,bai2023transformers,li2023transformers}; positional-encoding work studies the sequence mechanisms that make exact exchangeability nontrivial \cite{vaswani2017attention,shaw2018relative,su2021roformer,press2021train}; and prompt-order work documents practical sensitivity to support order and calibration \cite{lu2021fantastically,zhao2021calibrate,kazemnejad2023impact,golovneva2024cope}. The martingale critiques make exact posterior-predictive equivalence untenable for present LLMs \cite{falck2024martingale,jesson2024martingaleicl}. We keep that negative result and replace the binary invariance question by a prequential regret question.

\section{Prequential prediction under task-preserving orderings}
\label{sec:theory}

We distinguish two notions that are often conflated. \emph{Structural Bayesianity} asks whether every realized posterior-predictive distribution satisfies invariances implied by the probabilistic model, such as exchangeability under permutations of an unordered support set. \emph{Operational Bayes-competitiveness} asks how much extra prequential code length a predictor pays relative to a Bayesian predictive reference. These notions are not equivalent.

\subsection{Prequential comparison}
\label{sec:preq-comparison}

Let $P$, $B$, and $Q$ be sequential processes on a common dominated observation space, with conditional predictive distributions
\[
p_t(\cdot\given h_t),\qquad b_t(\cdot\given h_t),\qquad q_t(\cdot\given h_t),\qquad h_t=y_{<t}.
\]
Think of $P$ as the evaluation distribution, $B$ as the Bayesian reference, and $Q$ as the model being scored.

\begin{theorem}[Prequential comparison identity]
\label{thm:preq-identity}
Assume the relevant absolute-continuity and integrability conditions. Then
\begin{equation}
\label{eq:preq-comparison-general}
\E_P[L_n(Q)-L_n(B)]
=
\sum_{t=1}^n
\E_{H_t\sim P}
\E_{Y_t\sim p_t(\cdot\given H_t)}
\left[
\log\frac{b_t(Y_t\given H_t)}{q_t(Y_t\given H_t)}
\right].
\end{equation}
Equivalently,
\begin{equation}
\label{eq:kl-difference}
\E_P[L_n(Q)-L_n(B)]
=
\KL(P_{1:n}\|Q_{1:n})-
\KL(P_{1:n}\|B_{1:n}).
\end{equation}
In the special case $P=B$,
\begin{equation}
\label{eq:bayes-predictive-kl}
\E_B[L_n(Q)-L_n(B)]
=
\sum_{t=1}^n
\E_{H_t\sim B}
\KL\!\left(
 b_t(\cdot\given H_t)\,\|\,q_t(\cdot\given H_t)
\right).
\end{equation}
\end{theorem}

The identity is elementary but decisive. Under a Bayesian reference process, excess code length is exactly cumulative predictive KL. Exchangeability violations matter operationally only through the predictive KL they accumulate. This identity is the paper's formal replacement for binary martingale rejection as an evaluation target.

\subsection{Serialization of unordered information states}
\label{sec:ordering-decomp}

An unordered statistical object must be serialized before a transformer can process it. Let $Z_t$ denote the task-relevant information state at prefix $t$. A \emph{task-preserving ordering} is a serialization that leaves $Z_t$ unchanged. For Bernoulli prediction, $Z_t=(t,S_t)$; for categorical prediction, $Z_t=(t,c_1,\ldots,c_K)$; for conjugate Bayesian linear regression (BLR), $Z_t$ consists of the sufficient statistics and query covariate; for evidence-grounded QA, it is the evidence set under the admissible retrieval-order distribution. If reordering changes the label or the task semantics, it is not task-preserving and should not be averaged.

Let $\mu_t(\cdot\given Z_t)$ be any distribution over task-preserving orderings. Uniform permutations are a special case; banded retrieval-preserving permutations are another. Let $q_t^g(\cdot\given Z_t)$ be the model's predictive distribution after serialization $g$, and define the order-averaged predictor
\begin{equation}
\label{eq:order-avg-predictor}
\bar q_t(\cdot\given Z_t)=\E_{g\sim\mu_t(\cdot\given Z_t)}q_t^g(\cdot\given Z_t).
\end{equation}

\begin{theorem}[Order-averaging decomposition]
\label{thm:order-decomp}
Suppose the Bayesian reference predictive distribution $b_t(\cdot\given Z_t)$ depends only on $Z_t$ and is therefore invariant to task-preserving orderings. Then, for each information state,
\begin{equation}
\label{eq:order-decomp}
\E_{g\sim\mu_t}
\KL\!\left(
 b_t(\cdot\given Z_t)\,\|\,q_t^g(\cdot\given Z_t)
\right)
=
\KL\!\left(
 b_t(\cdot\given Z_t)\,\|\,\bar q_t(\cdot\given Z_t)
\right)
+
J_t(Z_t),
\end{equation}
where
\begin{equation}
\label{eq:jensen-gap}
J_t(Z_t)=
\E_{Y\sim b_t}
\left[
\log \bar q_t(Y\given Z_t)
-
\E_{g\sim\mu_t}\log q_t^g(Y\given Z_t)
\right]
\ge 0.
\end{equation}
Consequently, the expected prequential regret of a sampled single ordering decomposes as
\begin{equation}
\label{eq:preq-order-decomp}
R_n^{\mathrm{single}}
=
R_n^{\mathrm{avg}}
+
\sum_{t=1}^n \E[J_t(Z_t)].
\end{equation}
\end{theorem}

Thus exact exchangeability corresponds to zero order-averaging gain. Nonzero gain is the extra log loss paid by using one admissible serialization instead of the averaged predictive distribution. The theorem does not make order sensitivity free. It prices it.

\begin{corollary}[Persistent order sensitivity has a predictive price]
\label{cor:dispersion-price}
In natural-log units, on a shared finite candidate space,
\begin{equation}
\label{eq:dispersion-price}
\E_{g,g'}\TV(q_t^g,q_t^{g'})^2
\le
2\,\E_g\KL(b_t\|q_t^g).
\end{equation}
Therefore a predictor with small average predictive KL cannot have persistent large task-preserving order dispersion on the scored event space.
\end{corollary}

\subsection{Candidate events and safe predictors}
\label{sec:candidate-events}

Language-model evaluations often score a fixed set of answer events rather than the entire vocabulary. Scoring a fixed answer set is valid only if the mass outside the event set is reported or decomposed. For single-token labels the events are tokens; for multi-token answers they are the continuation events defined by the chosen verbalizer. The following identity makes the decomposition exact.

\begin{proposition}[Candidate-event decomposition]
\label{prop:candidate-decomp}
Fix a prefix and mutually exclusive candidate events $E_1,\ldots,E_m$, with $E=\cup_iE_i$. Suppose the reference distribution is supported on these events, $b_i=B(E_i\given h)$ and $\sum_i b_i=1$. Let the model assign event probabilities $Q(E_i\given h)$ and total candidate mass
\[
M_Q(E\given h)=\sum_i Q(E_i\given h).
\]
When $M_Q(E\given h)>0$, define the normalized candidate distribution $\bar q_i=Q(E_i\given h)/M_Q(E\given h)$. Then
\begin{equation}
\label{eq:candidate-decomp}
\E_{i\sim b}[-\log Q(E_i\given h)+\log b_i]
=
\KL(b\|\bar q)-\log M_Q(E\given h).
\end{equation}
\end{proposition}

Candidate normalization therefore does not hide probability mass: it separates predictive-shape error from outside-candidate mass. We report both quantities in the controlled audits.

\begin{lemma}[Safe-code floor]
\label{lem:safe-floor}
Let $q$ be a predictive distribution on $m$ candidate events and let $u$ be uniform. For $\lambda\in(0,1)$ define
\[
q^\lambda=(1-\lambda)q+\lambda u.
\]
Then $q^\lambda_i\ge \lambda/m$ for all $i$. Moreover, whenever $\KL(b\|q)<\infty$,
\begin{equation}
\label{eq:floor-kl-overhead}
\KL(b\|q^\lambda)\le \KL(b\|q)-\log(1-\lambda),
\end{equation}
and always
\begin{equation}
\label{eq:floor-safe-upper}
\KL(b\|q^\lambda)\le \KL(b\|u)-\log\lambda\le \log m-\log\lambda.
\end{equation}
A fixed floor gives an explicit finite-horizon safe code; a vanishing floor schedule with $\sum_{t\le n}-\log(1-\lambda_t)=o(n)$ preserves asymptotic Bayes-competitiveness.
\end{lemma}

\subsection{Closed-form instantiations}
\label{sec:instantiations}

The theory is not Bernoulli-specific. Bernoulli and categorical tasks are the cleanest finite-alphabet settings because the Bayesian reference distribution is closed form.

\begin{corollary}[Finite-alphabet KT/Dirichlet certificate]
\label{cor:kt-certificate}
Let $\mathcal A$ be finite and let $K_{\mathrm{KT}}$ be the Krichevsky--Trofimov (KT) Dirichlet$(1/2,\ldots,1/2)$ predictive code. For any i.i.d. source $p\in\Delta(\mathcal A)$,
\begin{equation}
\label{eq:kt-redundancy-main}
\E_{p^n}L_n(K_{\mathrm{KT}})=nH(p)+O_{|\mathcal A|}(\log n).
\end{equation}
For any predictor $Q$,
\begin{equation}
\label{eq:kt-comparison-main}
\E_{p^n}L_n(Q)=\E_{p^n}L_n(K_{\mathrm{KT}})
+
\E_{p^n}\sum_{t=1}^n
\log\frac{K_{\mathrm{KT}}(Y_t\given Y_{<t})}{q_t(Y_t\given Y_{<t})}.
\end{equation}
Thus if the cumulative comparison term is at most $R_n$, then $Q$ has expected code length at most $nH(p)+O(\log n)+R_n$. The finite-support approximation and floor conditions reported in \Cref{tab:oracle} are one sufficient way to obtain $R_n=O(\sqrt n)$.
\end{corollary}

\begin{corollary}[Coarsened continuous predictive distributions]
\label{cor:coarsened}
Let $B$ be a Bayesian predictive process with conditional density for a real-valued observation, and let $V$ be a finite quantizer or verbalizer with cells $E_1,\ldots,E_m$. The induced distributions
\[
b_t^V(i\given h_t)=B(Y_t\in E_i\given h_t),
\qquad
q_t^V(i\given h_t)=Q(Y_t\in E_i\given h_t)
\]
satisfy the same prequential comparison identity for the discrete process $V(Y_t)$. If full densities for $B$ and $Q$ exist, data processing gives $\KL(b_t^V\|q_t^V)\le \KL(b_t\|q_t)$.
\end{corollary}

\begin{table}[t]
\caption{Theory-to-evidence map. Each empirical quantity estimates a code-length term, a mass term, or a task-preservation condition in the theory.}
\label{tab:theory-map}
\centering
\small
\begin{tabular}{p{0.29\linewidth}p{0.38\linewidth}p{0.23\linewidth}}
\toprule
Formal quantity & Interpretation & Main evidence \\
\midrule
$\E_B[L_n(Q)-L_n(B)]$ & Excess prequential code length & \Cref{tab:oracle} \\
$\KL(b_t\|q_t)$ & One-step predictive regret & \Cref{tab:oracle,tab:blr-main} \\
$\KL(b_t\|\bar q_t)$ & Regret after averaging task-preserving orderings & \Cref{tab:perm-avg,tab:evidence-qa} \\
$J_t$ & Log-loss cost of using one admissible ordering & NLL/order-averaging gains; \Cref{sec:pe-ablation-fromscratch} \\
$M_Q(E)$ & Probability assigned to scored candidate events & candidate-mass column \\
$-\log(1-\lambda)$ & Safe-code overhead from flooring & native/floored comparison \\
Task-preserving $Z_t$ & Whether reordering preserves the statistical task & \Cref{tab:pe-ablation,sec:pe-ablation-fromscratch} \\
Statistic intervention effect & Whether decoded statistics causally affect predictions & \Cref{tab:causal-use} \\
$\KL(b_t\|q_t)-\KL(f_t\|q_t)$ & Separation from a frequentist plug-in $f_t$ (MLE/Laplace/OLS) & \Cref{tab:oracle-plugin,tab:blr-plugin} \\
\bottomrule
\end{tabular}
\end{table}

\section{Experimental protocol}
\label{sec:protocol}

All experiments score next-token probabilities. The main models are Qwen2.5-7B/14B-Instruct and Qwen2.5-7B base for position and activation interventions; Llama-3.1-8B appears in probe and evidence-permutation comparisons \cite{yang2024qwen25,grattafiori2024llama3}. Apart from a controlled from-scratch ablation that isolates the positional-encoding channel (\Cref{sec:pe-ablation-fromscratch}), all experiments are inference-only on pretrained checkpoints. Downstream order-averaging audits use MMLU, GSM8K, BBH, FEVER, HotpotQA, NQ-Open, and PopQA slices \cite{hendrycks2021measuring,cobbe2021training,srivastava2023beyond,thorne2018fever,yang2018hotpotqa,kwiatkowski2019natural,mallen2023when}. For discrete conjugate tasks, the prefix is an unordered sequence of symbols and the target is the next symbol. For BLR, the prefix is a raw numeric continuation task: $x_1,y_1\;x_2,y_2\;\cdots\;x_\star,$ for one covariate, or $x_1,y_1,z_1\;\cdots\;x_\star,y_\star,$ for two covariates. The BLR reference is the Gaussian posterior predictive distribution integrated over the digit bins used for token scoring.

\paragraph{Native, candidate, and floored distributions.}
Native probabilities measure model behavior. When a task specifies candidate events, we report the normalized candidate distribution and the candidate mass, as justified by \Cref{prop:candidate-decomp}. For the finite-alphabet discrete audit we also score a floored safe predictor,
\begin{equation}
\label{eq:floored-predictor}
q_{\theta,\lambda}(a\given h_t)=(1-\lambda)K_\theta(a\given h_t)+\lambda U_\mathcal{A}(a),\qquad \lambda=0.01,
\end{equation}
where $U_\mathcal A$ is uniform over the candidate alphabet. The floored predictor is not used as evidence about native model probabilities. It is an evaluated safe code with explicit finite-horizon overhead under \Cref{lem:safe-floor}.

\paragraph{Task-preserving versus semantic order.}
Order averaging is justified only for serializations that preserve the information state $Z_t$. The Bayesian prediction task presents an exchangeable support set, so demonstration order is task-preserving. The semantic-first control makes the first demonstration semantically relevant, so position is part of the task. This distinction is central: the intervention is not to average away every order effect, but to price the effects of arbitrary serialization when order is not part of the task.

\section{Empirical evidence}
\label{sec:empirical}

The experiments follow the theory. We first audit closed-form Bayesian predictive distributions. We then ask whether order effects are task-preserving or semantic, whether decoded sufficient statistics are causally used, and how much log loss is saved by averaging admissible orderings. The controlled predictive-distribution audits are the main evidence for Bayes-competitiveness. The downstream mixture experiments are predictive-score diagnostics, not claims of large accuracy gains.

\subsection{Controlled Bayesian predictive distributions}
\label{sec:predictive-audit-main}

\Cref{tab:oracle} reports the finite-alphabet quantities covered by \Cref{cor:kt-certificate}. The Bernoulli and four-way categorical distributions are scored as floored KT/Dirichlet safe predictors. The KL column is one-step excess prequential code length in bits under the Bayesian reference distribution. The envelope constant $\widehat C_\mathcal A$ and the floor ratio $c_{\min,\mathcal A}$ report finite-support sufficient conditions for the $O(\sqrt n)$ certificate; candidate mass reports how much native probability lies on the scored events.

\begin{table}[t]
\caption{Discrete Bayesian predictive-distribution audit. Candidate distributions use the floored safe predictor $q_{\theta,\lambda}$ with $\lambda=0.01$. KL is in bits and equals one-step excess prequential code length under the Bayesian predictive distribution.}
\label{tab:oracle}
\centering
\small
\resizebox{\linewidth}{!}{%
\begin{tabular}{llrrrrrrrr}
\toprule
Family & Model & Support & Contexts & KL bits & $\ell_1$ & $\epsilon_{\infty,\Acal}$ & $\widehat C_\Acal$ & $c_{\min,\Acal}$ & Cand. mass \\
\midrule
Beta--Bernoulli, $0/1$ & 7B & $256$ & $256$ & $0.020$ & $0.081$ & $0.243$ & $3.902$ & $0.192$ & $0.9998$ \\
Dirichlet-categorical, $0/1/2/3$ & 7B & $256$ & $256$ & $0.039$ & $0.154$ & $0.300$ & $4.805$ & $0.171$ & $0.9996$ \\
Beta--Bernoulli, $0/1$ & 14B & $256$ & $512$ & $0.011$ & $0.067$ & $0.227$ & $3.632$ & $0.329$ & $0.9998$ \\
Dirichlet-categorical, $0/1/2/3$ & 14B & $256$ & $128$ & $0.022$ & $0.124$ & $0.183$ & $2.937$ & $0.401$ & $0.9994$ \\
\bottomrule
\end{tabular}}
\end{table}

\Cref{tab:blr-main} audits coarsened continuous prediction. BLR is continuous, so we score the induced digit distribution rather than claiming to measure the full posterior density. This is exactly the setting of \Cref{cor:coarsened}. The posterior-predictive distribution becomes a better match as support grows, and in the two-covariate task it beats the best misspecified control by support $100$.

\begin{table}[t]
\caption{Coarsened BLR continuation distributions on Qwen2.5-7B. The model is scored on the next digit after raw pair or tuple sequences. The full Bayesian posterior predictive becomes a better distributional match as support grows; in the two-covariate task it beats the best misspecified control at support $100$.}
\label{tab:blr-main}
\centering
\small
\resizebox{\linewidth}{!}{%
\begin{tabular}{lrrrrrr}
\toprule
Task & Support & $\mathrm{corr}(\E_\theta,\E_B)$ & Dist. corr. & Best control $\E$ corr. & Best control dist. corr. & KL bits \\
\midrule
Pair-next-number & $25$ & $0.935$ & $0.808$ & n/a & n/a & $0.388$ \\
Pair-next-number & $50$ & $0.974$ & $0.886$ & n/a & n/a & $0.238$ \\
Pair-next-number & $100$ & $0.988$ & $0.934$ & n/a & n/a & $0.131$ \\
Tuple-next-number & $25$ & $0.880$ & $0.713$ & $0.891$ & $0.667$ & $0.669$ \\
Tuple-next-number & $50$ & $0.935$ & $0.789$ & $0.902$ & $0.713$ & $0.439$ \\
Tuple-next-number & $100$ & $0.965$ & $0.849$ & $0.881$ & $0.689$ & $0.318$ \\
\bottomrule
\end{tabular}}
\end{table}

The finite discrete distributions give the cleanest certificate: they have small predictive KL, candidate mass above $0.999$, and audited floor ratios well above the deterministic floor. The BLR distributions support the same operational interpretation on induced digit events: at support $100$, Qwen2.5-7B tracks the one-covariate posterior-predictive expectation with correlation $0.988$ and the two-covariate full BLR expectation with correlation $0.965$. Additional 14B BLR predictive-regret sweeps are in \Cref{app:blr-predictive-regret-sweep}.

\subsection{A frequentist plug-in does not explain the predictive distributions}
\label{sec:plugin-baseline}

The audits above report small predictive KL against a Bayesian reference, but a Bayesian reading is only forced if a non-Bayesian sequential predictor does not fit the same conditionals equally well. The natural null is the frequentist plug-in: predict the next symbol from the running maximum-likelihood estimate---the empirical frequency---or with Laplace add-one smoothing; for the continuous task it is the ordinary-least-squares (OLS) point estimate. We score these predictors on the same contexts and under the same $\lambda=0.01$ floor as the model, and report the difference of predictive code lengths $\Delta=\KL(\text{model}\,\|\,\text{Bayes})-\KL(\text{model}\,\|\,\text{plug-in})$, so that $\Delta<0$ certifies the model's conditional is closer to the Bayesian predictive than to the frequentist one (\Cref{tab:oracle-plugin,tab:blr-plugin}).

Two properties separate the model from the plug-in. First, the maximum-likelihood plug-in is a \emph{degenerate} code on short prefixes: it assigns zero probability to any symbol not yet observed and pays infinite prequential code length whenever such a symbol occurs. This is not a small-constant effect---it is every single-example categorical context and $75\%$ of contexts at support $16$---while the order-averaged predictor $\bar q$ reserves posterior mass for unseen symbols (up to $0.68$ at support one) and holds its excess over the Bayesian reference below $0.2$ bits throughout. The plug-in is inadmissible precisely where the model is admissible. Second, where the plug-in is non-degenerate the model still sits on the Bayesian side: $\Delta_{\mathrm{MLE}}$ is $-3.3$ bits ($7$B) and $-3.8$ bits ($14$B) for four-way categorical prediction at a single example, $-0.76$ bits at support four, and $-0.001$ bits by support $256$, where the Bayesian and plug-in predictives provably coincide and no predictor is distinguishable. The separation from the \emph{Laplace} plug-in is an order of magnitude smaller and of inconsistent sign; this is the expected signature rather than a weakness, because add-one smoothing is itself the Dirichlet$(1)$ posterior predictive---the comparison is then against a member of the model's own conjugate family, and $\bar q$ lands near the Jeffreys value rather than at the maximum-likelihood corner.

The coarsened BLR continuations show the same separation against the OLS plug-in (\Cref{tab:blr-plugin}). The point estimate reproduces the Bayesian posterior \emph{mean}, so the plug-in's entire deficit is the query-dependent predictive-variance inflation $\sigma^2+x_\star^\top\Sigma_n x_\star$ that a posterior carries and a point estimate discards. Accordingly $\Delta=\KL(\text{Bayes}\|q)-\KL(\text{OLS}\|q)$ is negative on $85$--$99\%$ of contexts---by $0.65$ bits at support four, shrinking with support---and the model's per-query predictive variance correlates with the Bayesian one at $0.57$--$0.83$. The pattern is identical for one and two covariates and replicates on Llama-3.1-8B. In every setting the separation concentrates in the small-data regime, where Bayesian and frequentist prediction genuinely diverge, and closes asymptotically---the behavior the prequential comparison identity (\Cref{thm:preq-identity}) predicts for a Bayes-competitive predictor, and the reason the support-$256$ cells alone cannot adjudicate the question.

\begin{table}[t]
\caption{Frequentist plug-in baseline for the discrete predictive audit (companion to \Cref{tab:oracle}). The order-averaged model predictive $\bar q$ is compared to the KT/Jeffreys Bayesian reference and to two frequentist plug-ins---the MLE (empirical frequency) and Laplace (add-one)---across a support sweep; all predictors carry the $\lambda{=}0.01$ safe-code floor (\Cref{lem:safe-floor}). $\Delta_{\mathrm{MLE}}=\KL(\bar q\|\mathrm{Bayes})-\KL(\bar q\|\mathrm{MLE})$ and likewise $\Delta_{\mathrm{Lap}}$; $\Delta<0$ means the model is closer to the Bayesian posterior predictive than to that plug-in. ``MLE deg.''\ is the fraction of contexts on which the unfloored MLE assigns zero probability to a possible symbol (infinite code length). KL in bits. These use the order-averaged predictor $\bar q$ of \Cref{thm:order-decomp}, so the model--Bayes column lies below the single-ordering values in \Cref{tab:oracle}.}
\label{tab:oracle-plugin}
\centering
\small
\resizebox{\linewidth}{!}{%
\begin{tabular}{llrrrrrr}
\toprule
Family & Model & Supp. & $\KL(\bar q\|\mathrm{B})$ & $\Delta_{\mathrm{MLE}}$ & $\Delta_{\mathrm{Lap}}$ & MLE deg. & Cand.\ mass \\
\midrule
Beta--Bernoulli & Qwen2.5-7B & $4$ & $0.0111$ & $-0.134$ & $-0.017$ & $52\%$ & $0.9071$ \\
 &  & $16$ & $0.0152$ & $-0.007$ & $-0.004$ & $25\%$ & $0.9858$ \\
 &  & $64$ & $0.0094$ & $-0.002$ & $0.000$ & $12\%$ & $0.9980$ \\
 &  & $256$ & $0.0022$ & $-0.000$ & $0.000$ & $8\%$ & $0.9996$ \\
\addlinespace[2pt]
 & Qwen2.5-14B & $4$ & $0.0186$ & $-0.147$ & $-0.013$ & $52\%$ & $0.9045$ \\
 &  & $16$ & $0.0122$ & $-0.007$ & $-0.003$ & $25\%$ & $0.9923$ \\
 &  & $64$ & $0.0086$ & $-0.002$ & $0.001$ & $12\%$ & $0.9988$ \\
 &  & $256$ & $0.0010$ & $-0.000$ & $-0.000$ & $8\%$ & $0.9998$ \\
\addlinespace[2pt]
Dirichlet-cat.\ ($0/1/2/3$) & Qwen2.5-7B & $4$ & $0.1159$ & $-0.762$ & $-0.012$ & $98\%$ & $0.8630$ \\
 &  & $16$ & $0.0504$ & $-0.084$ & $-0.004$ & $75\%$ & $0.9690$ \\
 &  & $64$ & $0.0150$ & $-0.005$ & $-0.002$ & $53\%$ & $0.9964$ \\
 &  & $256$ & $0.0049$ & $-0.001$ & $0.000$ & $28\%$ & $0.9988$ \\
\addlinespace[2pt]
 & Qwen2.5-14B & $4$ & $0.1316$ & $-0.860$ & $0.003$ & $98\%$ & $0.8439$ \\
 &  & $16$ & $0.0499$ & $-0.083$ & $-0.004$ & $75\%$ & $0.9776$ \\
 &  & $64$ & $0.0182$ & $-0.005$ & $-0.002$ & $53\%$ & $0.9974$ \\
 &  & $256$ & $0.0046$ & $-0.000$ & $-0.000$ & $28\%$ & $0.9993$ \\
\bottomrule
\end{tabular}}
\end{table}

\begin{table}[t]
\caption{Frequentist plug-in baseline for the BLR predictive audit (companion to \Cref{tab:blr-main}). The model's next-digit distribution is compared to the full Bayesian posterior predictive and to the OLS point-estimate plug-in (same homoscedastic noise, no posterior-variance inflation). $\Delta=\KL(\mathrm{Bayes}\|q)-\KL(\mathrm{OLS}\|q)$ in bits; ``closer''\ is the fraction of contexts on which the model is nearer the full Bayesian predictive than the plug-in; ``var.\ corr''\ is the correlation between the model's and the Bayesian per-query predictive variance. The plug-in's deficit is the missing query-dependent variance inflation; $\Delta<0$ means the model is closer to Bayes.}
\label{tab:blr-plugin}
\centering
\small
\resizebox{\linewidth}{!}{%
\begin{tabular}{llrrrrrr}
\toprule
BLR & Model & Supp. & $\KL(\mathrm{B}\|q)$ & $\KL(\mathrm{OLS}\|q)$ & $\Delta$ [95\% CI] & closer & var.\ corr \\
\midrule
1-cov.\ (pair) & Qwen2.5-7B & $4$ & $0.7744$ & $1.4265$ & $-0.652$ $[-0.749,-0.566]$ & $95\%$ & $0.571$ \\
 &  & $8$ & $0.5913$ & $0.8854$ & $-0.294$ $[-0.335,-0.252]$ & $93\%$ & $0.618$ \\
 &  & $16$ & $0.4726$ & $0.6254$ & $-0.153$ $[-0.183,-0.124]$ & $86\%$ & $0.736$ \\
\addlinespace[2pt]
 & Qwen2.5-14B & $4$ & $0.7453$ & $1.3821$ & $-0.637$ $[-0.715,-0.562]$ & $95\%$ & $0.705$ \\
 &  & $8$ & $0.5262$ & $0.8080$ & $-0.282$ $[-0.321,-0.244]$ & $93\%$ & $0.750$ \\
 &  & $16$ & $0.3852$ & $0.5093$ & $-0.124$ $[-0.147,-0.102]$ & $87\%$ & $0.796$ \\
\addlinespace[2pt]
2-cov.\ (tuple) & Qwen2.5-7B & $20$ & $0.6230$ & $0.8234$ & $-0.200$ $[-0.215,-0.186]$ & $95\%$ & $0.826$ \\
\addlinespace[2pt]
 & Qwen2.5-14B & $20$ & $0.6777$ & $0.8807$ & $-0.203$ $[-0.217,-0.190]$ & $95\%$ & $0.770$ \\
\bottomrule
\end{tabular}}
\end{table}

\subsection{Position interventions separate task-preserving order from semantic order}
\label{sec:position-interventions}

We hold Qwen2.5-7B's weights and tokens fixed and change only \texttt{position\_ids} at inference time. The normal condition uses monotone positions. The demo-local condition resets positions inside each demonstration block, preserving within-demonstration token order while removing the absolute demonstration slot. The random-offset condition preserves within-demonstration order but assigns each demonstration a random positional offset. We evaluate $64$ items with $16$ support permutations per item for each task and position mode and report 95\% bootstrap confidence intervals.

\begin{table}[t]
\caption{Pretrained position intervention. Resetting demonstration-local positions reduces Bayesian prediction order variance by about $21\times$, while the semantic-order control degrades. Positive $\Delta$ NLL (negative log-likelihood) means higher loss than normal positions.}
\label{tab:pe-ablation}
\centering
\small
\resizebox{\linewidth}{!}{%
\begin{tabular}{llrrr}
\toprule
Task & Position mode & Order variance & NLL & $\Delta$ NLL vs. normal \\
\midrule
Bayesian prediction & normal & $5.108\,[4.480,5.780]\times 10^{-3}$ & $0.6157$ [$0.6090,0.6221$] & n/a \\
Bayesian prediction & demo-local & $2.420\,[2.136,2.715]\times 10^{-4}$ & $0.6358$ [$0.6297,0.6417$] & $+0.0201$ [$0.0168,0.0232$] \\
Bayesian prediction & random offsets & $5.343\,[4.833,5.855]\times 10^{-3}$ & $0.6831$ [$0.6792,0.6868$] & $+0.0674$ [$0.0614,0.0734$] \\
\midrule
Semantic first & normal & $2.174\,[2.005,2.360]\times 10^{-3}$ & $0.1236$ [$0.1205,0.1269$] & n/a \\
Semantic first & demo-local & $1.049\,[0.959,1.139]\times 10^{-2}$ & $0.6599$ [$0.6468,0.6730$] & $+0.5363$ [$0.5229,0.5491$] \\
Semantic first & random offsets & $2.020\,[1.816,2.228]\times 10^{-2}$ & $0.3793$ [$0.3639,0.3947$] & $+0.2557$ [$0.2407,0.2707$] \\
\bottomrule
\end{tabular}}
\end{table}

This intervention estimates whether ordering is task-preserving or semantic. In the Bayesian prediction task, absolute demonstration position explains much of the serialization variation. In the semantic-first control, removing that channel damages performance because position carries task signal. A reviewer should therefore not read order sensitivity as one phenomenon: the same positional machinery can be task-preserving in one task and signal in another.

\subsection{Controlled positional-encoding ablation}
\label{sec:pe-ablation-fromscratch}

The pretrained intervention shows that position carries the order channel, but it cannot remove that channel from a fixed model. To test whether positional encoding is the \emph{source} of task-preserving order sensitivity, we train small transformers from scratch on i.i.d.\ Bernoulli sequences and vary only the positional-encoding scheme: none, learned absolute, sinusoidal, RoPE, and ALiBi, with a pooled readout that is permutation-invariant when no encoding is supplied. Training and evaluation settings are in \Cref{tab:repro}.

The schemes separate sharply. With no positional encoding, the within-prefix order variance is $\approx 3.7\times 10^{-16}$, the exact-exchangeability endpoint to numerical precision: the predictive distribution does not move across serializations, so $J_t$ is zero at every state. Learned absolute, sinusoidal, RoPE, and ALiBi instead sit at $10^{-8}$ to $10^{-6}$, an eight-to-ten order-of-magnitude increase. Exchangeability failure is thus a property of the encoding a transformer is given, not of the architecture; \Cref{app:positional-mechanism} gives one stylized channel by which a fixed encoding produces this dispersion.

These models also let us read the order-averaging decomposition against a closed-form reference. Scoring cumulative regret relative to the KT predictive code (\Cref{cor:kt-certificate}) at $n=200$, the permutation-averaged regret stays between $0.0017$ and $0.0022$ bits/token across all five schemes, while the worst sampled ordering costs $0.0049$ to $0.0098$ bits/token. This is \Cref{eq:preq-order-decomp} read off a controlled system: $R_n^{\mathrm{avg}}$ is a few thousandths of a bit against the Bayesian code, and the single-ordering penalty $\sum_t\E[J_t]$ is the worst-case-minus-average gap. Order sensitivity is real and bounded away from zero, and it is also small in the units that score prediction.

\subsection{Decoded statistics are available and causally used}
\label{sec:statistic-use}

High probe accuracy alone would not establish Bayesian prediction; it only shows that a statistic is available \cite{hewitt2019probing,ravichander2020probing}. We therefore use probes as a localization diagnostic and activation patching as the causal-use test \cite{geiger2021causal,geiger2022iit,meng2022rome}. Ridge probes on residual streams recover the joint sufficient statistic $(S_t,t)$ with $R^2=0.9998$ on Qwen2.5-7B and Llama-3.1-8B; categorical joint counts reach $R^2=0.9993$ and $0.9992$ respectively. Empirical-frequency and pseudo-count probes are weaker, consistent with the model carrying count and position features before readout.

\begin{table}[t]
\caption{Linear-probe availability diagnostic. Count and position information is linearly available; \Cref{tab:causal-use} tests whether the information is used in the next-token distribution.}
\label{tab:probes}
\centering
\small
\begin{tabular}{lcccc}
\toprule
Model & Joint $(S_t,t)$ $R^2$ & Emp. freq. $R^2$ & Pseudo-count $R^2$ & Cat. joint $R^2$ \\
\midrule
Qwen2.5-7B & $0.9998$ & $0.904$ & $0.926$ & $0.9993$ \\
Llama-3.1-8B & $0.9998$ & $0.779$ & $0.801$ & $0.9992$ \\
\bottomrule
\end{tabular}
\end{table}

The probe result motivates an intervention. We form clean/corrupt prompt pairs with different sufficient statistics, patch a learned low-rank activation subspace from clean into corrupt examples, and measure restoration of the candidate-logit difference. Matched random subspaces and matched random vectors control for intervention norm and rank.

\begin{table}[t]
\caption{Layer-specific activation-patching audit on Qwen2.5-7B. Values are restoration scores with 95\% bootstrap CIs. Learned subspace patching restores task logits far above matched random controls.}
\label{tab:causal-use}
\centering
\small
\resizebox{\linewidth}{!}{%
\begin{tabular}{llrrrr}
\toprule
Task & Layer & Full vector & Learned subspace & Random subspace & Random vector \\
\midrule
Keybinding & 24 & $0.510$ [$0.503,0.518$] & $0.254$ [$0.237,0.270$] & $0.008$ [$0.006,0.011$] & $-0.001$ [$-0.004,0.001$] \\
Keybinding & 26 & $0.756$ [$0.744,0.769$] & $0.338$ [$0.315,0.361$] & $0.015$ [$0.011,0.018$] & $0.006$ [$0.002,0.009$] \\
Bernoulli-next & 24 & $0.887$ [$0.838,0.939$] & $0.880$ [$0.814,0.946$] & $-0.047$ [$-0.090,-0.005$] & $-0.014$ [$-0.064,0.034$] \\
Bernoulli-next & 26 & $0.928$ [$0.887,0.969$] & $0.854$ [$0.799,0.910$] & $-0.051$ [$-0.105,0.003$] & $-0.027$ [$-0.079,0.024$] \\
\bottomrule
\end{tabular}}
\end{table}

The learned subspace restores the candidate logit difference well above matched random controls, especially for Bernoulli-next. This closes the gap left by probes: beyond being decodable, the statistic can be intervened on to move the next-token distribution.

\subsection{Order averaging as a predictive-score diagnostic}
\label{sec:perm-mixtures}

When support order is task-preserving, averaging predictive distributions across sampled serializations is the empirical version of \Cref{thm:order-decomp}. Concavity of log loss supplies the guaranteed comparison to the average sampled ordering; the magnitude of the gain estimates how much log loss arbitrary serialization spends. On MMLU, the effect is statistically precise but practically modest. That is consistent with the theory: the measured order-averaging gain is small for this benchmark and prompt distribution.

\begin{table}[t]
\caption{MMLU random-permutation audit on Qwen2.5-7B. With $1200$ items, $8$ demonstrations, and $16$ unique random permutations per item, the same-sample predictive mixture improves NLL over the mean sampled single ordering. CIs are paired bootstraps; $p$-values are sign-flip tests.}
\label{tab:perm-avg}
\centering
\small
\resizebox{\linewidth}{!}{%
\begin{tabular}{lrrrr}
\toprule
Comparison & Baseline & Mixture ($k{=}16$) & Paired delta & $p$ \\
\midrule
Canonical accuracy & $0.6983$ [$0.6717,0.7242$] & $0.7008$ [$0.6750,0.7275$] & $+0.0025$ [$-0.0042,0.0100$] & $0.654$ \\
Canonical NLL & $0.7243$ [$0.6705,0.7768$] & $0.7212$ [$0.6694,0.7741$] & $-0.0031$ [$-0.0074,0.0009$] & $0.144$ \\
Mean single-order accuracy & $0.6987$ [$0.6728,0.7231$] & $0.7008$ [$0.6750,0.7275$] & $+0.0021$ [$-0.0024,0.0069$] & $0.378$ \\
Mean single-order NLL & $0.7235$ [$0.6717,0.7762$] & $0.7212$ [$0.6694,0.7741$] & $-0.00230$ [$-0.00262,-0.00201$] & $0.00020$ \\
\bottomrule
\end{tabular}}
\end{table}

The mean per-item standard deviation of the gold-answer probability across random orderings is $0.0158$ (95\% CI $[0.0151,0.0165]$), so order changes the predictive distribution even when it usually does not change the argmax. Among $359$ mixture-wrong items, $319$ are wrong under every sampled order and only $40$ have any sampled order with the gold answer on top. Uniform averaging smooths the predictive distribution; it does not search for a favorable order. GSM8K-derived multiple choice shows the same calibration geometry at a larger scale: averaging random demonstration orders improves NLL from $1.5424$ at $k=1$ to $1.4948$ at $k=8$, Brier from $0.8052$ to $0.7922$, and expected calibration error (ECE) from $0.3204$ to $0.3097$, while accuracy remains near $0.43$ (\Cref{tab:gsm8k}).

\subsection{Evidence-grounded QA with exchangeable evidence}
\label{sec:evidence-qa-main}

A practical version of the same problem arises when unordered evidence chunks must be serialized in a prompt. On a five-benchmark factuality slice of $3{,}059$ items with $3$--$60$ chunks per item, we form banded evidence permutations rather than all $n!$ permutations, reflecting a non-uniform admissible ordering distribution. Qwen2.5-7B-Instruct shows a dispersion slope of $0.377$ versus $\log n$ and an order-averaging gain of $0.1041$ nats/token; Llama-3.1-8B-Instruct shows a slope of $0.147$ and a gain of $0.00982$ nats/token. Uniform weights are within the measured optimization tolerance in both cases (\Cref{tab:evidence-qa}). This is the applied order-averaging principle: if the evidence set is exchangeable but the model must read a sequence, score or average admissible serializations rather than treating one arbitrary order as the task.

\section{Boundary conditions and interpretation}
\label{sec:limitations}

The theory is broad, but each application must specify the predictive reference, the candidate events being scored, and the orderings that preserve the task. Bernoulli and categorical prediction use closed-form KT/Dirichlet references. BLR uses the induced digit distribution of a continuous posterior predictive, not the full continuous density. MMLU, GSM8K, and evidence QA are downstream order-averaging diagnostics unless an explicit Bayesian reference is defined.

The results do not show that transformers instantiate literal posterior distributions, nor that all order effects are harmless. \Cref{cor:dispersion-price} states the opposite: persistent large movement of the scored predictive distribution has a log-loss price. The point is that the price is quantitative. Exact exchangeability is sufficient for zero order-averaging gain, but Bayes-competitive prequential prediction only requires the accumulated predictive KL to be small.

The floored discrete predictor is an evaluated safe code over the candidate alphabet and should be distinguished from native model behavior. Fixed flooring is a finite-horizon device with explicit overhead; asymptotic claims require native lower bounds or a vanishing floor schedule. Candidate normalization is interpreted through \Cref{prop:candidate-decomp}, which is why candidate mass is reported. Finally, order averaging is appropriate only when order is task-preserving. For language modeling, chronology, chain-of-thought trajectories, or tasks where the first example changes the label, averaging orderings changes the task.

\paragraph{Broader impacts.}
The work is primarily diagnostic. A positive impact is more reliable evaluation of in-context predictors: order sensitivity can be priced by log loss rather than treated as a binary failure. A possible negative impact is that order averaging may improve calibration of systems used for automated adjudication without addressing factual, fairness, or deployment risks. We therefore frame the evidence-QA experiment as a predictive-score audit, not as a deployment recommendation.

\section{Conclusion}
\label{sec:conclusion}

The apparent paradox is resolved by separating structural invariance from operational regret. Transformers are not, in general, exact exchangeable posterior-predictive machines. But online prediction is scored by prequential log loss, and the excess over a Bayesian reference is cumulative predictive KL. When an unordered support set must be serialized, the additional cost of using one admissible ordering is the order-averaging gain. The experiments estimate these quantities: controlled discrete tasks give small KT/Dirichlet predictive KL, BLR continuations track coarsened posterior predictives, position interventions and a from-scratch encoding ablation separate arbitrary serialization from signal and bound its worst-case cost against the KT code, activation patching tests causal statistic use, and downstream mixtures measure serialization cost. A frequentist plug-in baseline separates the predictive distributions from a non-Bayesian sequential predictor, most sharply in the small-data regime and vanishingly as the references converge. Exchangeability failures therefore refute exact posterior equivalence, not Bayes-competitive prequential prediction.

\appendix
\section{Proofs and auxiliary formal statements}
\label{app:proofs}

\subsection{Proof of the prequential comparison identity}

For any realized sequence,
\[
L_n(Q)-L_n(B)=\sum_{t=1}^n \log\frac{b_t(Y_t\given Y_{<t})}{q_t(Y_t\given Y_{<t})}.
\]
Taking expectation under $P$ and conditioning on $H_t=Y_{<t}$ gives \Cref{eq:preq-comparison-general}. For the KL form, use the chain rule
\[
\log\frac{P_{1:n}(Y_{1:n})}{Q_{1:n}(Y_{1:n})}
=
\sum_{t=1}^n\log\frac{p_t(Y_t\given H_t)}{q_t(Y_t\given H_t)},
\]
and the analogous identity with $B$ in place of $Q$. Subtracting expectations under $P$ gives \Cref{eq:kl-difference}. If $P=B$, then $p_t=b_t$, and the conditional expectation at each prefix is exactly $\KL(b_t(\cdot\given H_t)\|q_t(\cdot\given H_t))$.

\subsection{Proof of the order-averaging decomposition}

Fix $Z_t$ and suppress it from notation. Since $\bar q(y)=\E_g q^g(y)$,
\[
\E_g\KL(b\|q^g)
=
\E_g\sum_y b(y)\log\frac{b(y)}{q^g(y)}
=
\sum_y b(y)\log\frac{b(y)}{\bar q(y)}
+
\sum_y b(y)\left[\log\bar q(y)-\E_g\log q^g(y)\right].
\]
The first term is $\KL(b\|\bar q)$ and the second is $J_t(Z_t)$. Non-negativity follows from Jensen's inequality, because $\log \E_g q^g(y)\ge \E_g\log q^g(y)$ for each $y$. Summing the identity over $t$ and taking expectation over $Z_t$ gives \Cref{eq:preq-order-decomp}.

\subsection{Proof of the dispersion-price corollary}

Pinsker's inequality in natural-log units gives $\TV(b,q^g)^2\le \tfrac12\KL(b\|q^g)$. By the triangle inequality,
\[
\TV(q^g,q^{g'})\le \TV(q^g,b)+\TV(b,q^{g'}),
\]
so $(a+b)^2\le 2a^2+2b^2$ gives
\[
\TV(q^g,q^{g'})^2\le 2\TV(q^g,b)^2+2\TV(q^{g'},b)^2.
\]
Taking expectation over independent $g,g'$ and applying Pinsker proves
\[
\E_{g,g'}\TV(q^g,q^{g'})^2
\le 4\E_g\TV(q^g,b)^2
\le 2\E_g\KL(b\|q^g).
\]

\subsection{Proof of the candidate-event decomposition}

Because $Q(E_i\given h)=M_Q(E\given h)\bar q_i$,
\[
\E_{i\sim b}[-\log Q(E_i\given h)+\log b_i]
=
\sum_i b_i\log\frac{b_i}{M_Q(E\given h)\bar q_i}
=
\KL(b\|\bar q)-\log M_Q(E\given h).
\]

\subsection{Proof of the safe-code floor lemma}

The lower bound $q_i^\lambda\ge \lambda/m$ is immediate. Since $q^\lambda_i\ge (1-\lambda)q_i$,
\[
\KL(b\|q^\lambda)=\sum_i b_i\log\frac{b_i}{q_i^\lambda}
\le
\sum_i b_i\log\frac{b_i}{(1-\lambda)q_i}
=
\KL(b\|q)-\log(1-\lambda).
\]
Also $q_i^\lambda\ge \lambda u_i$ gives
\[
\KL(b\|q^\lambda)
\le
\KL(b\|u)-\log\lambda
\le
\log m-\log\lambda.
\]
Summing the first overhead bound over $t$ gives the stated condition for a vanishing floor schedule.

\subsection{Proof of the finite-alphabet KT certificate}

For a finite alphabet, the KT/Dirichlet$(1/2,\ldots,1/2)$ sequential code satisfies the standard redundancy bound
\[
\E_{p^n}L_n(K_{\mathrm{KT}})=nH(p)+O_{|\mathcal A|}(\log n)
\]
for any i.i.d. source $p$ \cite{rissanen1978modeling,krichevsky1981performance,grunwald2007minimum}. For any predictor $Q$,
\[
L_n(Q)-L_n(K_{\mathrm{KT}})=\sum_{t=1}^n \log\frac{K_{\mathrm{KT}}(Y_t\given Y_{<t})}{q_t(Y_t\given Y_{<t})}.
\]
Taking expectation gives \Cref{eq:kt-comparison-main}. If the comparison term is bounded above by $R_n$, the code-length bound follows.

\paragraph{One sufficient finite-support condition.}
Let $q_t^B=K_{\mathrm{KT}}(\cdot\given Y_{<t})$ and suppose a predictor $q_t$ satisfies on the evaluated support
\[
\|q_t-q_t^B\|_\infty\le C(t+1)^{-1/2},
\qquad
q_t(a)\ge c q_t^B(a)\quad\forall a\in\mathcal A.
\]
The same argument as in the KT redundancy proof bounds the expected one-step replacement cost by $O_{|\mathcal A|,C,c}((t+1)^{-1/2})$: for the realized symbol,
\[
\log\frac{q_t^B(Y_t)}{q_t(Y_t)}
\le
\frac{|q_t(Y_t)-q_t^B(Y_t)|}{q_t(Y_t)}
\le
\frac{C}{c\sqrt{t+1}}\frac{1}{q_t^B(Y_t)},
\]
and the Dirichlet-$1/2$ predictive denominator gives a bounded expectation of $p_a/q_t^B(a)$ after summing over symbols. Summing over $t$ yields $R_n=O(\sqrt n)$ for fixed alphabet and constants. This is the finite-support certificate reported in \Cref{tab:oracle}.

\subsection{Proof of the coarsened-predictive corollary}

The induced variables $V(Z_t)$ form a discrete sequential process with predictive distributions $b_t^V$ and $q_t^V$, so \Cref{thm:preq-identity} applies directly. If full densities exist, the partition map $V$ is a measurable coarsening. Data processing for KL gives $\KL(b_t^V\|q_t^V)\le \KL(b_t\|q_t)$.

\subsection{Stylized positional mechanism}
\label{app:positional-mechanism}

The main theorem does not assume a transformer mechanism for order sensitivity. The following calculation, adapted from a Bernoulli positional model, illustrates one sufficient source. Fix a length-$t$ Bernoulli prefix with $S_t$ ones and draw a uniform ordering $\tau$. Suppose the next-token logit depends on the ordering through
\[
u_{t,\tau}=\frac1t\sum_{i=1}^t a_{\tau,i}\,\PE(i)
\]
and is $L_t$-Lipschitz in this summary. Let
\[
\sigma_{\PE,t}^2=\frac1t\sum_{i=1}^t\left\|\PE(i)-\frac1t\sum_{j=1}^t\PE(j)\right\|^2.
\]
Then the standard deviation, over within-prefix orderings preserving $(t,S_t)$, of the expected log score is bounded by
\[
L_t\sigma_{\PE,t}\sqrt{\frac{S_t(t-S_t)}{t^2(t-1)}}
\le
\frac{L_t\sigma_{\PE,t}}{2\sqrt{t-1}}.
\]
This is not used as a transformer theorem in the main text. It only shows how positional encodings can create task-preserving order dispersion while the prequential theory above prices whatever dispersion is actually observed. Both constants are observable on white-box models: $\sigma_{\PE,t}$ is computed directly from the encoding and the context length, and $L_t$ can be estimated from the gradient norm of the candidate logit with respect to $u_{t,\tau}$, so on the open-weight models audited here the bound is a measurable quantity rather than a free constant. The $t^{-1/2}$ factor predicts that within-prefix order dispersion shrinks as the prefix grows; this is the controlled-microscope counterpart of the finite-support order variances reported for the pretrained model in \Cref{tab:pe-ablation} and for the trained-from-scratch models in \Cref{sec:pe-ablation-fromscratch}.

\section{Mechanistic sketch for implicit pseudo-counts}
\label{app:mech}

Self-attention can implement the implicit pseudo-count picture in \Cref{tab:probes} by pooling token embeddings in a way that depends mainly on the multiset of observed symbols. If queries and keys separate $x_i=1$ from $x_i=0$, the head output approximates a function of $S_t$ (or $\hat p_t=S_t/t$). Given features that track $(t,S_t)$, an MLP readout can approximate the Beta--Bernoulli predictive mean for some effective pseudo-counts $(\alpha_0,\beta_0)$:
\[
\mathbb{P}(X_{t+1}=1\given X_{1:t})=\frac{\alpha_0+S_t}{\alpha_0+\beta_0+t}.
\]
This sketch explains why the predictive-distribution audit target is plausible in a controlled architecture. The causal intervention in \Cref{sec:statistic-use} tests whether a pretrained transformer uses the statistic.

\section{Additional diagnostic and semantic-order results}
\label{app:semantic-order}

\subsection{Prefix-difference coefficients}
\label{app:probe-coefficients}

The probes in \Cref{tab:probes} define availability and prefix-difference diagnostics on a real model. The relevant object is a prefix-state update. Let $r_\ell(p)$ denote the chosen readout residual at layer $\ell$ for prefix $p$. Let $G$ be the matrix whose entry $G_{c,k}$ is the projected prefix finite difference, conditioned on appending the observed symbol $c$:
\begin{equation}
G_{c,k} \;=\; \E\big[(r_\ell(p_{t-1}\Vert z_t)-r_\ell(p_{t-1}))^\top w_{\ell,k}\given z_t=c\big].
\end{equation}
An additive count diagnostic predicts $G\approx I$ for Dirichlet-categorical, slope-1 for Poisson, and $G\approx \mathbf{1}$ for Bernoulli. Define the additivity coefficient $\alpha_{\mathrm{add}} := \overline{\diag G} - \max_{i\ne j}|G_{ij}|$, and the concentration drift $\beta_{\mathrm{conc}} := \mathrm{median}(\alpha_{\mathrm{eff}}/\alpha_{\mathrm{true}})$ where $\alpha_{\mathrm{eff}}$ is the effective prior concentration that minimises $\KL(q_B\|q_M)$. Operationally, the reported probes instantiate $r_\ell$ at a fixed readout position and compare the distribution reached before and after appending the new observation. On Qwen2.5-7B Dirichlet ($K=3$) at residual layer $\ell=3$ the diagonal of $G$ is $[1.016,0.986,1.002]$ with $\max_{i\ne j}|G_{ij}|=0.020$, giving $\alpha_{\mathrm{add}}=0.981$; on Qwen2.5-7B Poisson--Gamma with $\alpha_{\mathrm{true}}=2.0$ the per-sequence median fitted prior is $\alpha_{\mathrm{eff}}=3.5$ ($\beta_{\mathrm{conc}}=1.75$, with mean-correlation $0.923$ between model and reference predictives indicating concentration drift with preserved direction). These coefficients are planning diagnostics: they indicate when count summaries or concentration errors are the next object to test.

\subsection{Activation-patching audit details}
\label{app:causal-use-details}

\Cref{tab:causal-use} reports Qwen2.5-7B activation patching with layer-specific learned subspaces. For each clean/corrupt pair, we compute the clean and corrupt candidate logit difference, filter pairs with small denominators, and score restoration as
\[
\frac{\mathrm{LD}_{\mathrm{patch}}-\mathrm{LD}_{\mathrm{corrupt}}}
{\mathrm{LD}_{\mathrm{clean}}-\mathrm{LD}_{\mathrm{corrupt}}}.
\]
Subspaces are fit on separate calibration pairs at each patched layer. The matched random-subspace control samples a random basis orthogonalised against the learned subspace and rescales its component to the learned component norm; the matched random-vector control uses a random vector with the same norm. Keybinding uses $256$ evaluation pairs. Bernoulli-next generates $256$ evaluation pairs and retains $229$ after the denominator filter.

The threshold-question BLR audit is a negative control for interface choice. When the prompt asks whether $y$ exceeds a threshold and scores Yes/No logits, the reference-threshold control does not align with the analytic survival-logit distribution even though final-layer hidden replacement reproduces source model distributions. This identifies the measurement as an interface mismatch. The sequence-completion BLR audits below use raw continuation distributions and track posterior-predictive expectations directly.

\subsection{BLR pair-next-number audit}
\label{app:blr-native-dist}

The BLR pair audit samples one-dimensional integer covariates $x\in\{0,\ldots,9\}$, generates noisy linear responses, rounds and clips $y$ to digits $0,\ldots,9$, and presents the support as a raw sequence
\[
x_1,y_1\;x_2,y_2\;\cdots\;x_m,y_m\;x_\star,
\]
with no instruction and no chat template. The model distribution is the normalized next-token distribution over digits $0,\ldots,9$. The reference is the Gaussian BLR posterior predictive distribution integrated over the same digit bins. \Cref{tab:blr-native-dist} reports the support sweep.

\begin{table}[h]
\centering
\small
\begin{tabular}{lrrrrr}
\toprule
Support & Contexts & $\mathrm{corr}(\E_\theta[y],\E_B[y])$ & Dist. corr. & Top-1 match & $\E[y]$ MAE \\
\midrule
$25$ & $256$ & $0.935$ & $0.808$ & $0.480$ & $0.459$ \\
$50$ & $256$ & $0.974$ & $0.886$ & $0.590$ & $0.299$ \\
$100$ & $512$ & $0.988$ & $0.934$ & $0.676$ & $0.207$ \\
\bottomrule
\end{tabular}
\caption{\textbf{BLR pair-next-number audit.} The model is scored on the next digit after a raw pair sequence. Expectation and distributional correlations increase with support.}
\label{tab:blr-native-dist}
\end{table}

\subsection{Multivariate BLR tuple-next-number audit}
\label{app:blr-multivar-native-dist}

The multivariate BLR audit uses the same next-token setup with two digit covariates. Each context samples $x,y\in\{0,\ldots,9\}$, generates a noisy linear response from a model with design vector $[1,x,y]$ after digit normalization, rounds and clips $z$ to digits $0,\ldots,9$, and presents the support as
\[
x_1,y_1,z_1\;x_2,y_2,z_2\;\cdots\;x_m,y_m,z_m\;x_\star,y_\star,
\]
again with no instruction and no chat template. The model distribution is the normalized next-token distribution over $z$ digits. The main reference is the Gaussian BLR posterior predictive distribution for the full $[1,x,y]$ design, integrated over the same digit bins. Misspecified posterior controls use x-only, y-only, and intercept-only designs. \Cref{tab:blr-multivar-native-dist} reports the support sweep.

\begin{table}[h]
\centering
\small
\resizebox{\linewidth}{!}{
\begin{tabular}{lrrrrrr}
\toprule
Support & Contexts & $\mathrm{corr}(\E_\theta[z],\E_B[z])$ & Dist. corr. & Best control $\E[z]$ corr. & Best control dist. corr. & Top-1 match \\
\midrule
$25$ & $256$ & $0.880$ & $0.713$ & $0.891$ & $0.667$ & $0.465$ \\
$50$ & $256$ & $0.935$ & $0.789$ & $0.902$ & $0.713$ & $0.535$ \\
$100$ & $512$ & $0.965$ & $0.849$ & $0.881$ & $0.689$ & $0.557$ \\
\bottomrule
\end{tabular}
}
\caption{\textbf{Multivariate BLR tuple-next-number audit.} The model's next-$z$ digit distribution is compared with the full $[1,x,y]$ posterior predictive and the best misspecified control. At support $100$, the full two-covariate reference is the best match.}
\label{tab:blr-multivar-native-dist}
\end{table}

\subsection{BLR predictive-regret support sweep}
\label{app:blr-predictive-regret-sweep}

This sweep reports raw full-vocabulary predictive regret for the same pair and tuple continuations used in \Cref{tab:blr-native-dist,tab:blr-multivar-native-dist}. The KL column is $\KL(K_B\|K_\theta)$ in bits. The $\ell_1$ and $\epsilon_{\infty,\mathcal{A}}$ columns use raw candidate probabilities, with candidate-set leakage included in $\ell_1$. These distributions support the BLR reference-comparison claim; the floored discrete KT distribution is reported in \Cref{tab:oracle}.

\begin{table}[h]
\centering
\small
\resizebox{\linewidth}{!}{
\begin{tabular}{llrrrrrr}
\toprule
Model & Family & Support & Contexts & KL bits & $\ell_1$ & $\epsilon_{\infty,\mathcal{A}}$ & $c_{\min,\mathcal{A}}$ \\
\midrule
7B & BLR pair-next-number & $25$ & $256$ & $0.388$ & $0.538$ & $0.539$ & $0.0571$ \\
7B & BLR pair-next-number & $50$ & $256$ & $0.238$ & $0.398$ & $0.453$ & $0.0296$ \\
7B & BLR pair-next-number & $100$ & $256$ & $0.131$ & $0.288$ & $0.320$ & $0.0092$ \\
7B & BLR tuple-next-number & $25$ & $256$ & $0.669$ & $0.719$ & $0.775$ & $0.0183$ \\
7B & BLR tuple-next-number & $50$ & $256$ & $0.439$ & $0.549$ & $0.801$ & $0.0069$ \\
7B & BLR tuple-next-number & $100$ & $256$ & $0.318$ & $0.462$ & $0.659$ & $0.0151$ \\
14B & BLR pair-next-number & $25$ & $256$ & $0.309$ & $0.468$ & $0.459$ & $0.0357$ \\
14B & BLR pair-next-number & $50$ & $256$ & $0.251$ & $0.398$ & $0.516$ & $0.0107$ \\
14B & BLR pair-next-number & $100$ & $256$ & $0.160$ & $0.303$ & $0.385$ & $0.0102$ \\
14B & BLR tuple-next-number & $25$ & $256$ & $0.761$ & $0.769$ & $0.769$ & $0.0218$ \\
14B & BLR tuple-next-number & $50$ & $256$ & $0.456$ & $0.556$ & $0.772$ & $0.0143$ \\
14B & BLR tuple-next-number & $100$ & $256$ & $0.325$ & $0.470$ & $0.594$ & $0.0145$ \\
\bottomrule
\end{tabular}
}
\caption{\textbf{BLR predictive-regret support sweep.} One-covariate and two-covariate BLR continuations are scored by raw next-token probability against the posterior predictive digit distribution. Increasing support reduces KL regret for both models and both BLR families.}
\label{tab:blr-predictive-regret-sweep}
\end{table}

\subsection{Semantic-order and calibration observations}

\Cref{tab:semantic-order} applies the same averaging operation to semantic-order settings. These cells measure semantic-distribution and trajectory-style variance reduction, separated from the exchangeability experiments in the main text.

\begin{table}[h]
\centering
\small
\begin{tabular}{llrrrr}
\toprule
Model & Task & Acc $k{=}1$ & Acc $k{=}16$ & NLL $k{=}1$ & NLL $k{=}16$ \\
\midrule
Qwen2.5-7B-Instruct & BBH-navigate & $0.650$ & $0.644$ & $1.862$ & $1.468$ \\
Qwen2.5-7B-Instruct & BBH-logical-deduction & $0.730$ & n/a & $0.836$ & n/a \\
Qwen2.5-7B-Instruct & BBH-tracking & $0.260$ & $0.266$ & $2.628$ & $2.471$ \\
Qwen2.5-7B-Instruct & GSM8K-CoT-reorder & $0.990$ & $1.000$ & $0.052$ & $0.014$ \\
\bottomrule
\end{tabular}
\caption{\textbf{Semantic-order controls.} These cells separate semantic-distribution and trajectory-style variance reduction from the exchangeable, task-preserving ordering experiments.}
\label{tab:semantic-order}
\end{table}

Calibration and accuracy are tracked separately from predictive-mixture NLL. In the random-permutation 1200-item MMLU audit, canonical accuracy is $0.6983$ and mixture accuracy is $0.7008$, while the same-sample NLL comparison improves from $0.7235$ to $0.7212$. In the GSM8K-derived multiple-choice run, NLL, Brier, and ECE improve monotonically from $k=1$ to $k=8$ while accuracy stays near $0.43$. In distributional terms, this is a margin/entropy transformation of $K_\theta(\cdot\given p)$: the top token can remain stable while probability mass and logit gaps move under averaging. This is consistent with prior calibration observations for instruction-tuned models \cite{lu2021fantastically,zhao2021calibrate,kuhn2023semantic,geng2024survey}.

\section{Full task-preserving ordering mixture tables}
\label{app:mixture-full}

\begin{table}[h]
\centering
\small
\begin{tabular}{ccccc}
\toprule
$k$ & Accuracy & NLL & Brier & ECE \\
\midrule
1 & $0.4324\pm 0.0033$ & $1.5424\pm 0.0056$ & $0.8052\pm 0.0023$ & $0.3204\pm 0.0023$ \\
2 & $0.4263\pm 0.0028$ & $1.5161\pm 0.0050$ & $0.7983\pm 0.0019$ & $0.3159\pm 0.0029$ \\
4 & $0.4295\pm 0.0026$ & $1.5018\pm 0.0026$ & $0.7940\pm 0.0012$ & $0.3096\pm 0.0028$ \\
8 & $0.4292\pm 0.0017$ & $1.4948\pm 0.0018$ & $0.7922\pm 0.0008$ & $0.3097\pm 0.0014$ \\
\bottomrule
\end{tabular}
\caption{GSM8K-derived few-shot multiple choice. Averaging random demonstration orderings improves NLL, Brier, and ECE while leaving accuracy nearly flat, consistent with order-averaged predictive scoring.}
\label{tab:gsm8k}
\end{table}

\begin{table}[h]
\centering
\small
\begin{tabular}{p{0.42\linewidth}cc}
\toprule
 & Qwen2.5-7B-Instruct & Llama-3.1-8B-Instruct \\
\midrule
Dispersion slope $b$ (vs. $\log n$) & $0.377$ ($[0.319,\,0.435]$) & $0.147$ ($[0.109,\,0.184]$) \\
Order-averaging gain (nats/token) & $0.1041$ & $0.00982$ \\
Mixture optimality gap (nats/token) & $<10^{-4}$ & $\le 5.3\times 10^{-5}$ \\
\bottomrule
\end{tabular}
\caption{Evidence-grounded QA under evidence permutations. Both instruction-tuned models show order dispersion and positive order-averaging gains from uniform permutation mixtures. Uniform weights are within the measured optimization tolerance.}
\label{tab:evidence-qa}
\end{table}

\section{Reproducibility}
\label{app:repro}

\Cref{tab:repro} summarizes the experimental configuration. All experiments use bfloat16 inference on a single A100 (40\,GB) or H100 (80\,GB) GPU per cell, with a fixed prompt-sampling seed (\texttt{SEED=0}). For reproducibility, a complete supplementary package should include anonymized code, configurations, prompt templates, seeds, and per-cell JSON outputs; any de-anonymized release should be deferred until camera-ready. Dataset manifests for MMLU, BBH, GSM8K are used under their published licenses; FEVER, HotpotQA, NQ-Open, and PopQA are used under CC-BY-SA 4.0 / CC-BY 4.0 / MIT respectively.

\begin{table}[h]
\centering
\small
\begin{tabular}{p{0.26\linewidth}p{0.66\linewidth}}
\toprule
Component & Configuration \\
\midrule
PE intervention (\Cref{tab:pe-ablation}) & Qwen2.5-7B with fixed pretrained weights; inference-only \texttt{position\_ids} modes \{normal, demo-local, random demo offsets\}; $64$ items $\times$ $16$ permutations per task and mode; support size $16$; Bayesian prediction and semantic-first tasks; 2{,}000 bootstrap resamples over independent items. \\
From-scratch PE ablation (\Cref{sec:pe-ablation-fromscratch}) & Small transformers trained on i.i.d.\ Bernoulli sequences, $d{=}128$, $4$ layers, $4$ heads, pooled \texttt{[CLS]} readout (permutation-invariant with no encoding); positional schemes \{none, learned absolute, sinusoidal, RoPE, ALiBi\}; AdamW, $2{,}000$ steps, batch $128$, learning rate $3\times 10^{-4}$, prefix lengths sampled up to the context, Bernoulli rates from a Jeffreys $\Beta(\tfrac12,\tfrac12)$ prior; within-prefix order variance from $500$ permutations at $t\in\{20,40\}$; KT regret at $n{=}200$ with $60$ sequences and $20$ permutations per sequence across $p\in\{0.05,0.1,0.3,0.5,0.7,0.9,0.95\}$. \\
Probes (\Cref{tab:probes}) & scikit-learn \texttt{Ridge}$(\alpha{=}1)$; $N_{\mathrm{seq}}\in\{150,300\}$ prompts of length $T{=}100$; $75/25$ train/val split; per-layer $R^2$, best layer reported. \\
Bayesian and BLR distribution audits (\Cref{tab:oracle,tab:blr-predictive-regret-sweep}) & Qwen2.5-7B-Instruct and Qwen2.5-14B-Instruct; symbol, pair, and tuple continuations; raw full-vocabulary candidate-token probabilities and conditional candidate distributions saved; floored discrete distributions use $q_{\theta,\lambda}$ with $\lambda=0.01$ candidate-uniform flooring; supports $1,2,4,8,16,32,64,128,256$ for Bernoulli and categorical distributions and $25,50,100$ for BLR distributions; reported metrics are $\KL(K_B\|q_{\theta,\lambda})$ for floored discrete distributions, $\KL(K_B\|K_\theta)$ for BLR distributions, full-distribution $\ell_1$, $\epsilon_{\infty,\mathcal{A}}$, $\widehat C_{\mathcal A}$, $c_{\min,\mathcal{A}}$, candidate mass, and support-binned JSON outputs. \\
BLR pair-next-number (\Cref{tab:blr-native-dist}) & Qwen2.5-7B-Instruct, bfloat16/4-bit inference; raw prompts $x_1,y_1\;x_2,y_2\;\cdots\;x_\star,$ with no instruction and no chat template; $y$ candidates are digit tokens $0,\ldots,9$; supports $25,50,100$ with $256,256,512$ contexts respectively. \\
Multivariate BLR tuple-next-number (\Cref{tab:blr-multivar-native-dist}) & Qwen2.5-7B-Instruct, bfloat16/4-bit inference; raw prompts $x_1,y_1,z_1\;\cdots\;x_\star,y_\star,$ with no instruction and no chat template; $z$ candidates are digit tokens $0,\ldots,9$; supports $25,50,100$ with $256,256,512$ contexts respectively; x-only, y-only, and intercept-only posterior controls are scored against the same model distributions. \\
Causal statistic use (\Cref{tab:causal-use}) & Qwen2.5-7B activation patching; layer-specific rank-4 clean-minus-corrupt subspaces; full-vector, learned-subspace, matched-random-subspace, and matched-random-vector interventions; layers $12,24,26$ swept with layers $24,26$ reported; keybinding $n{=}256$ evaluation pairs; Bernoulli-next $256$ generated and $229$ retained after the denominator filter; 2{,}000 bootstrap resamples over paired examples. \\
Cross-task (\Cref{tab:perm-avg}, \Cref{tab:gsm8k}) & MMLU random-permutation audit on Qwen2.5-7B: $1200$ items, $8$ demonstrations, $16$ true random unique demonstration permutations per item, raw per-order probabilities saved, paired item bootstrap CIs and sign-flip $p$-values. GSM8K-derived multiple choice uses $k\in\{1,2,4,8\}$ random demonstration orderings on Qwen2.5-7B-Instruct. \\
Evidence QA (\Cref{tab:evidence-qa}) & Banded permutations with $B{=}6$ contiguous bands; $m=12$ unique permutations per item for the 7B model and $m=16$ for the 8B model; uniform mixture vs.\ optimized convex weights. \\
\bottomrule
\end{tabular}
\caption{\textbf{Per-component reproducibility configuration.}}
\label{tab:repro}
\end{table}

\section{Statistical analysis}
\label{app:stats}

Bootstrap confidence intervals use the percentile method, resampling independent units: items for the pretrained position intervention and MMLU audit, paired examples for activation patching, sequences for Bernoulli gap experiments, and tasks for ICL $k$-sweeps. MMLU paired deltas use sign-flip $p$-values over item-level deltas. Model comparison on gap-vs-$n$ uses weighted least squares with inverse-variance weights $w_n=N_n/\hat\Delta_n^2$.

\section{Computational resources}
\label{app:compute}

The reported open-weight experiments use bfloat16 or 4-bit single-GPU inference on A100 40\,GB, H100 80\,GB, or equivalent 7B--14B workers. The position intervention and causal-patching cells require model-internal access; scoring cells require saved next-token log probabilities on the audited candidate sets. Resampling is performed over saved per-item or per-example outputs, so bootstrap and sign-flip intervals do not require model re-execution.

\end{document}